\documentclass{article}

\usepackage{microtype}
\usepackage{graphicx}
\usepackage{booktabs} 

\usepackage{hyperref}

\usepackage{bm}                             
\usepackage{amsmath,amssymb}                       
\usepackage{dsfont}
\usepackage{soul}
\usepackage{verbatim}
\usepackage{caption}
\usepackage{subcaption}
\usepackage{graphicx}
\usepackage{graphbox}
\usepackage{makecell}
\usepackage{multirow}
\usepackage{wrapfig}
\usepackage[table]{xcolor}
\usepackage[T1]{fontenc}

\definecolor{Gray}{gray}{0.9}

\DeclareMathOperator*{\argmax}{arg\,max}

\renewcommand{\comment}[1]{}

\newcommand{\stdv}[1]{\scalebox{.55}{~$\pm$~#1}}

\usepackage[accepted]{icml2021}
\icmltitlerunning{Self-Improved Retrosynthetic Planning}

\begin{document}

\twocolumn[
\icmltitle{Self-Improved Retrosynthetic Planning}

\icmlsetsymbol{equal}{*}

\begin{icmlauthorlist}
\icmlauthor{Junsu Kim}{kaist}
\icmlauthor{Sungsoo Ahn}{mbzuai}
\icmlauthor{Hankook Lee}{kaist}
\icmlauthor{Jinwoo Shin}{kaist}
\end{icmlauthorlist}

\icmlaffiliation{kaist}{Korea Advanced Institute of Science and Technology (KAIST)}
\icmlaffiliation{mbzuai}{Mohamed bin Zayed University of Artificial Intelligence (MBZUAI)}

\icmlcorrespondingauthor{Junsu Kim}{junsu.kim@kaist.ac.kr}

\icmlkeywords{Machine Learning, ICML}

\vskip 0.3in
]

\printAffiliationsAndNotice{}  

\begin{abstract}
Retrosynthetic planning is a fundamental problem in chemistry for finding a pathway of reactions to synthesize a target molecule. Recently, search algorithms have shown promising results for solving this problem by using deep neural networks (DNNs) to expand their candidate solutions, i.e., adding new reactions to reaction pathways. However, the existing works on this line are suboptimal; the retrosynthetic planning problem requires the reaction pathways to be (a) represented by real-world reactions and (b) executable using ``building block'' molecules, yet the DNNs expand reaction pathways without fully incorporating such requirements. 
Motivated by this, 
we propose an end-to-end framework for directly training the DNNs towards generating reaction pathways with the desirable properties. Our main idea is based on a self-improving procedure that trains the model to imitate successful trajectories found by itself. We also propose a novel reaction augmentation scheme based on a forward reaction model. Our experiments demonstrate that our scheme significantly improves the success rate of solving the retrosynthetic problem from 86.84\% to 96.32\% while maintaining the performance of DNN for predicting valid reactions. 
\end{abstract}

\section{Introduction}
\begin{figure}[t]
\vspace{0.1in}
\begin{subfigure}{\columnwidth}
\centering
\includegraphics[width=1\linewidth]{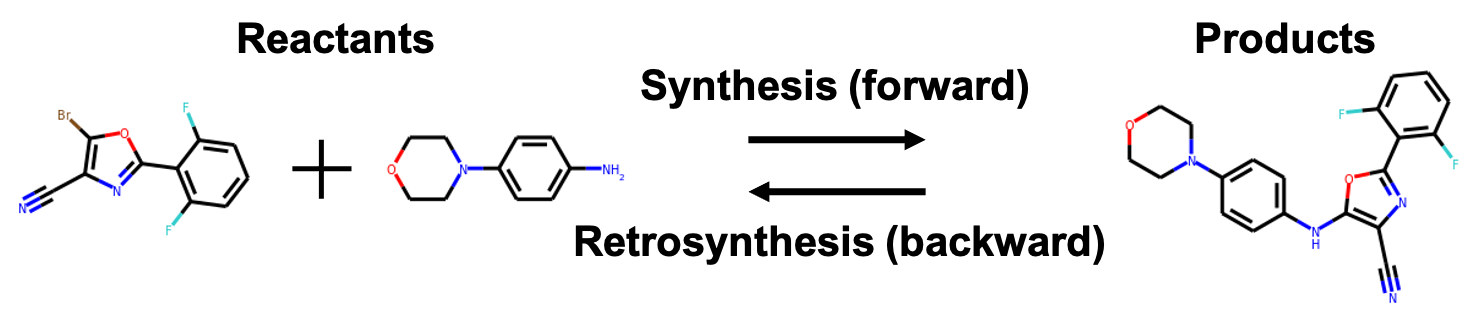}
\caption{Chemical reaction}
\end{subfigure}
\vspace{0.1in}
\begin{subfigure}{\columnwidth}
\centering
\includegraphics[width=0.92\linewidth]{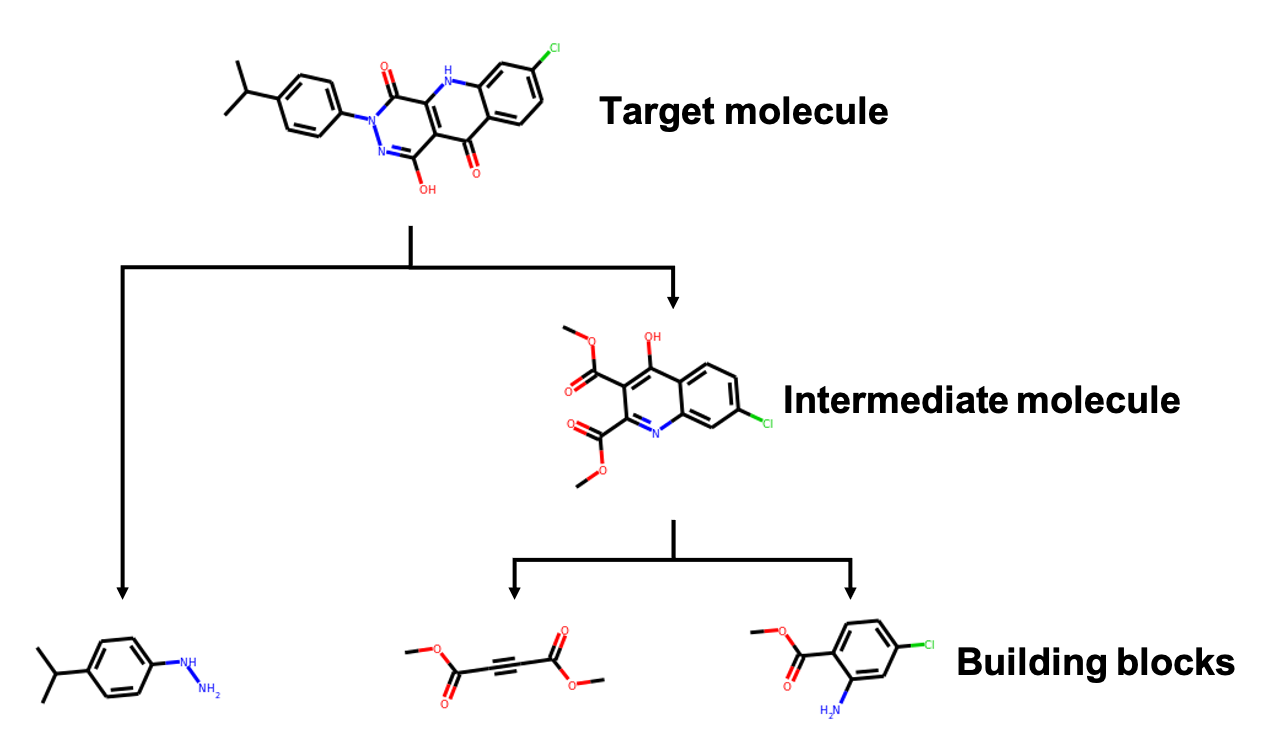}
\caption{Retrosynthetic planning}
\end{subfigure}
\caption{Illustration of (a) synthesis (forward) and retrosynthesis (backward) with respect to a chemical reaction and (b) outcome of retrosynthetic planning given the target molecule {2-(2,6-difluorophenyl)-5-(4-morpholin-4-ylanilino)-1,3-oxazole-4-carbonitrile}. Given a target molecule, a retrosynthetic planning algorithm aims at finding a reaction pathway ending up in the building block molecules.}
\end{figure}
\begin{figure*}[t]
\vspace{0.1in}
\centering
\includegraphics[width=0.92\linewidth]{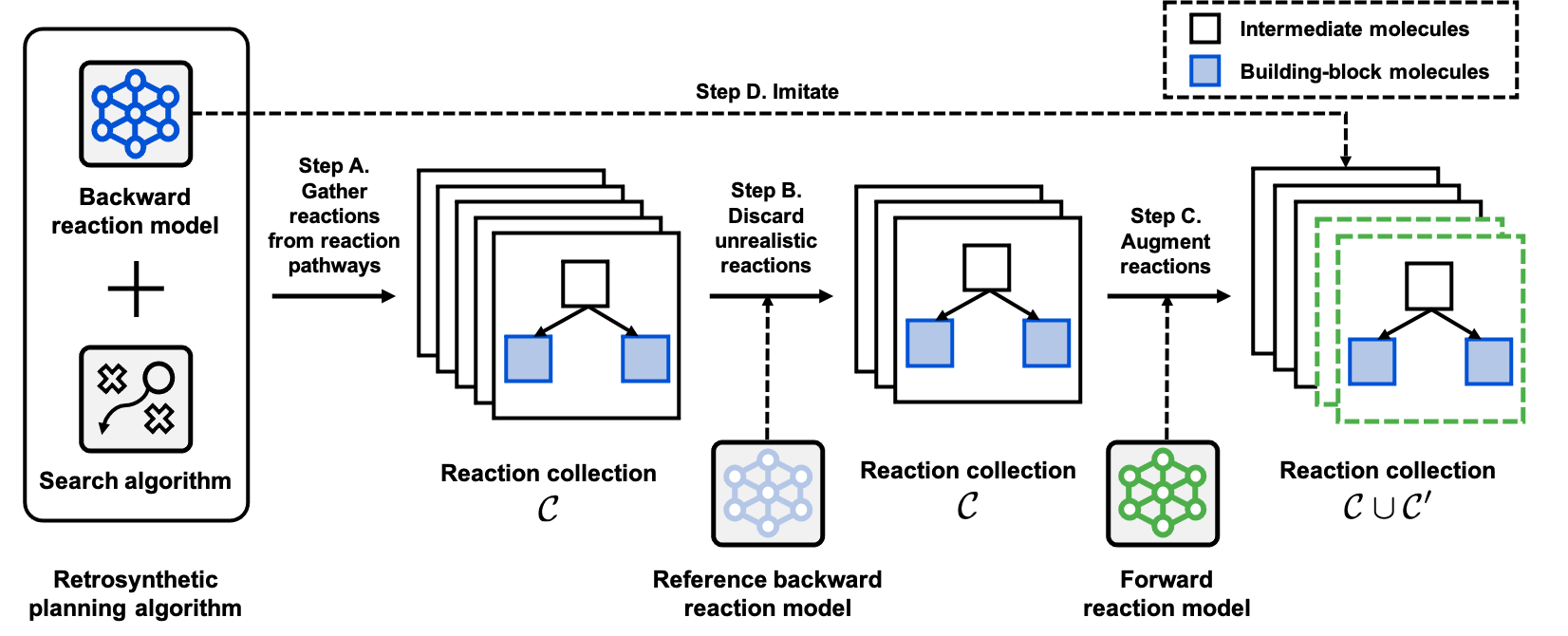}
\caption{Illustration of our framework. Our framework iterates the four-step procedure. In step A, we gather reactions from the reaction pathways that are generated via a search algorithm combined with a backward reaction model to form a collection of reactions $\mathcal{C}$. In step B, we discard unrealistic reactions in the collection $\mathcal{C}$ using a reference backward model. In step C, we generate a set of reactions $\mathcal{C}'$ from augmenting the reactions in the collection $\mathcal{C}$ using a forward reaction model. In step D, we train the backward reaction model to imitate reactions in the collection $\mathcal{C} \cup \mathcal{C}'$.}
\label{fig:concept}
\end{figure*}

To synthesize a novel molecule, chemists require executing a pathway of reactions starting from a set of known or commercially available building block molecules. Hence, discovering such a reaction pathway for a target molecule is crucial in important applications such as drug discovery \citep{hughes2011principles} and material design \citep{yan2018non}. 


To tackle this problem,
retrosynthetic planning \citep{corey1991logic} finds a series of chemically valid reactions starting from the target molecule until reaching the building block molecules in a backward and recursive manner. The main challenge of retrosynthetic planning is twofold: (a) finding an accurate single-step retrosynthetic model that predicts a single reaction of a given product and (b) designing an efficient search algorithm for a reaction pathway starting from the set of building block molecule. 

Especially, recent works have proposed deep neural networks (DNNs) as attractive models for single-step retrosynthesis. Using the existing real-world reaction datasets \citep{lowe2012extraction}, they train (in a supervised manner) and evaluate DNNs 
to predict a reactant-set from a given product. To be specific, existing works use DNNs to predict the reactant-set based on applying a known reaction template to the product \citep{segler2017neural, dai2019retrosynthesis}, generating each reactant from scratch \citep{liu2017retrosynthetic, karpov2019transformer, zheng2019predicting}, or modifying the product using atom-wise and bond-wise operations \citep{shi2020graph, somnath2020learning, yan2020retroxpert}. 

On the other hand, researchers also {have} developed efficient search algorithms for retrosynthetic planning based on the DNN-based single-step retrosynthetic models. Their main idea is to represent retrosynthetic planning as a sequential decision making problem and apply tree search algorithms such as Monte Carlo tree search \citep{segler2018planning}, proof number search \citep{kishimoto2019depth}, and A* search \citep{chen2020retro}. 

Intriguingly, most of the existing DNN-based retrosynthetic planning frameworks are not end-to-end. Namely, the performance of a retrosynthetic planning algorithm can be evaluated by (a) whether if the algorithm proposes reaction pathways representing reactions existing in real-world and (b) the success rate of finding such reaction pathways starting from the set of building block molecules. Since existing frameworks optimize DNN-based single-step retrosynthetic models and search algorithms for (a) and (b) separately, they may have suboptimal performance. 

\textbf{Contribution.} In this paper, we propose a new end-to-end framework for 
retrosynthetic planning based on training the DNN-based single-step retrosynthetic model toward maximizing the performance of retrosynthetic planning. 
We train DNNs for maximizing the success rate of search algorithms in addition to representing the inverse of real-world reactions. While our framework can be simply implemented on top of existing frameworks for retrosynthetic planning, we empirically observe that our end-to-end training of DNN leads to surprisingly large performance gains. 

To train the single-step retrosynthetic model for maximizing the success rate of search algorithms, we introduce a self-improving procedure that trains the model to imitate successful trajectories found by itself combined with the search algorithm. To train the model to generate realistic reaction pathways, we additionally introduce a likelihood-based criterion for filtering out samples used in the self-improving algorithm. Finally, to improve the generalization ability of the single-step retrosynthetic model, we propose a novel augmentation scheme based on modifying reactions using a forward reaction model. We provide an overall illustration of our framework in Figure \ref{fig:concept}.

To demonstrate the effectiveness of our framework, we conduct experiments based on the USPTO database \cite{lowe2012extraction}. Thanks to imitating reactions that are realistic and executable from building block molecules, our framework significantly improves the success rate of solving the retrosynthetic problem from 86.84\% to 96.32\%. Moreover, the reduced average time for planning demonstrates the efficacy of our framework. The average length and cost of searched pathways, which is a metric to measure the quality, also decreased than other baselines. In our ablation studies, we show the effectiveness of each component in our framework.

Our work 
reduces the gap between the widely used supervised learning of single-step retrosynthetic models and the goal of retrosynthetic planning. We believe that our work would guide new interesting directions in the future by bridging this gap.

\section{Preliminary}
\subsection{Problem Setup}
The task of retrosynthetic planning is to search for a set of chemical reactions $\tau=\{R_{i}\}_{i=1}^{N}$, i.e., a reaction pathway, required for synthesizing a target molecule $t$. Each reaction $R=(m, \mathcal{R})$ is represented by a pair of a product $m$ and a reactant-set $\mathcal{R} = \{r_{j}\}_{j=1}^{M}$.\footnote{To simplify the problem, we omit other conditions for describing a reaction, e.g., reagents. However, they can be incorporated into our framework with relatively small modifications.} Furthermore, the reaction pathways are desired to satisfy the following conditions:
\begin{itemize}
    \item[A.] The reactions should correspond to a realistic pair of a product and a reactant-set, i.e., any reaction $R$ should be executable in the real world. 
    \item[B.] Any reactant $r$ in a reaction $R \in \tau$ should be either a member of building block molecules $\mathcal{I}$ or a product of another reaction $R^{\prime} \in \tau$.
\end{itemize}

For training and evaluating retrosynthetic planning algorithms, we assume having access to datasets $\mathcal{D}_{\mathtt{target}}$ and $\mathcal{D}_{\mathtt{reaction}}$ consisting of real-world target molecules and reactions, respectively.






\subsection{Forward and Backward Reaction Models}
We consider retrosynthetic planning based on deep neural networks (DNNs) for representing a \textit{backward reaction model} $p_{b}(\mathcal{R} | m; \theta_{b})$  with parameter $\theta_{b}$, i.e., a single-step restrosynthetic model. In addition, we consider a \textit{forward reaction model} $p_{f}(m | \mathcal{R}; \theta_{f})$ with parameter $\theta_{f}$. Such a choice allows the reaction models to flexibly incorporate the complex chemical knowledge based on the expressive power of DNNs.

Researchers have developed various ways of modeling forward reaction \citep{jin2017predicting, bradshaw2018generative, schwaller2018found, do2019graph, schwaller2019molecular}, and backward reaction  \citep{segler2017neural, dai2019retrosynthesis, liu2017retrosynthetic, karpov2019transformer, zheng2019predicting, shi2020graph, somnath2020learning, yan2020retroxpert} using DNNs. They are mainly categorized into either template-based or template-free approach depending on their reliance on the reaction templates, i.e., subgraph patterns describing how the chemical reaction occurs among reactants.



In this work, we consider template-based reaction models represented by a multi-layered perceptron (MLP), prioritizing templates to apply in the given templates list \cite{segler2017neural}. To be specific, the MLP is trained to predict plausible templates to apply, taking Morgan fingerprint \citep{rogers2010extended}, a fixed-size vector representation of a set of molecules, as an input. 


\subsection{Search Algorithms}
To solve retrosynthetic planning with respect to the exponentially large space of reaction pathways, it is crucial to use an efficient \textit{search algorithm} $\mathcal{A}$. Existing search algorithms \citep{segler2018planning, kishimoto2019depth, chen2020retro} build reaction pathways by updating them in a backward direction, i.e., adding reactions for synthesizing a product in the intermediate reaction pathway. They typically use a backward reaction model trained on a real-world dataset to propose a reaction pathway consisting of realistic reactions. The data structure and the protocol for updating the intermediate reaction pathways are specific to search algorithms, e.g., Monte Carlo tree search \citep{segler2017neural} and proof number search \citep{kishimoto2019depth}. 

To be specific, \citet{segler2018planning} design a search tree where each node represents a set of molecules and expand the tree by balancing the selection of high-value nodes and unexplored nodes. The value of each node is estimated based on a Monte Carlo rollout of backward reaction models. Furthermore, \citet{kishimoto2019depth} and \citet{chen2020retro} employ AND-OR search trees to represent retrosynthetic pathways, where OR and AND nodes represent molecules and reactions, respectively. The most promising node to expand during planning is selected using human designed heuristics \citep{kishimoto2019depth} or a DNN trained on an offline dataset \citep{chen2020retro}.

In this work, we consider the recently proposed \textsc{Retro*} \cite{chen2020retro} for traversing the space of reaction pathways since it has demonstrated strong performance. The \textsc{Retro*} algorithm mimics the A* algorithm by performing the best-first search based on the cost of the current path and the estimated cost to the goal. They consider the cost of the current path as the sum of reaction costs, and the estimated cost to the goal is computed from the value function, which is parameterized by a neural network trained using an existing dataset of reactions. 
Additionally, \citet{chen2020retro} also introduce \textsc{Retro*-0}, which does not utilize the value function for expansion that makes a tradeoff between its performance and the expense of using an additional DNN for representing the value function.

\section{Self-Improved Retrosynthetic Planning} 
\subsection{Overview of Self-Improved Retrosynthetic Planning}
In this section, we introduce a new framework for retrosynthetic planning based on a self-improved model adaptation procedure. Similar to prior works, our framework aims to find reaction pathways by running a search algorithm $\mathcal{A}$ using reactions suggested by a backward reaction model $p_{b}(m | \mathcal{R}; \theta_{b})$ with parameter $\theta_{b}$. However, our framework differs from the existing works by adapting the backward reaction model towards improving the performance of search algorithms; existing works use backward reaction models that are agnostic to the choice of search algorithm and may yield suboptimal performance.

At a high level, our framework trains the backward reaction model $p_{b}(m | \mathcal{R}; \theta_{b})$ to maximize the likelihood of generated reaction pathways from itself combined with search algorithm $\mathcal{A}$. To improve the quality of reactions from the reaction pathways used for imitation learning, we introduce a \textit{reference backward reaction model} $p_{b}(m | \mathcal{R}; \bar{\theta}_{b})$ using the same architecture as the original backward reaction model. It is trained on a real-world reaction dataset $\mathcal{D}_{\mathtt{reaction}}$ to let its likelihood determine whether a reaction resembles reactions existing in the real world. 
Furthermore, we propose a novel reaction augmentation scheme based on a forward reaction model $p_{f}(m | \mathcal{R}; \theta_{f})$ to improve the diversity of reactions used for imitation learning. 
To further clarify our objective, note that we aim to optimize both the success rate of planning and the synthesis route cost. Following the prior work \citep{chen2020retro}, we consider $-\sum_{(m, \mathcal{R}) \in \tau}\log p_{b}(\mathcal{R} | m ; \bar{\theta}_{b})$ as the synthesis route cost: accumulation of negative log-likelihoods of reactions under the reference backward model.

Our framework repeats the following steps:
\begin{itemize}
    \item \textbf{Step A.} Generate a set of reaction pathways based on combining a search algorithm with the backward reaction model. Gather reactions from the reaction pathways to form a collection of reactions $\mathcal{C}$.
    \item \textbf{Step B.} Discard any reactions from the collection $\mathcal{C}$ that are determined to be unrealistic using a reference backward reaction model.
    \item \textbf{Step C.} Generate a set of reactions $\mathcal{C}^{\prime}$ from augmenting the reactions in the collection $\mathcal{C}$ by replacing the corresponding product using a forward reaction model. 
    \item \textbf{Step D.} Train the backward reaction model by maximizing the log-likelihood of the reactions in $\mathcal{C}\cup\mathcal{C}^{\prime}$, i.e., maximize $\sum_{(m, \mathcal{R}) \in \mathcal{C}\cup\mathcal{C}^{\prime}}\log p_{b}(\mathcal{R} | m; \theta_{b})$.
\end{itemize}

\begin{algorithm}[tb]
   \caption{Self-Improved Retrosynthetic Planning}
   \label{alg:algo}
\begin{algorithmic}
\STATE {\bfseries Input:} backward reaction model $p_{b}$, forward reaction model $p_{f}$, retrosynthetic planning algorithm $\mathcal{A}$, target molecule dataset $\mathcal{D}_{\mathtt{target}}$, reaction dataset $\mathcal{D}_{\mathtt{reaction}}$, and filtering thresholds $\epsilon, \epsilon_{\mathtt{aug}}$. 
\STATE Maximize $\sum_{(m, \mathcal{R})\in \mathcal{D}_{\mathtt{reaction}}} p_{b}(\mathcal{R} | m; \theta_{b})$ over $\theta_{b}$.
\STATE Maximize $\sum_{(m, \mathcal{R})\in \mathcal{D}_{\mathtt{reaction}}} p_{b}(\mathcal{R} | m; \bar{\theta}_{b})$ over $\bar{\theta}_{b}$.
\STATE Maximize $\sum_{(m, \mathcal{R})\in \mathcal{D}_{\mathtt{reaction}}} p_{f}(m | \mathcal{R}; \theta_{f})$ over $\theta_{f}$.
\FOR{$i=1,\ldots, I$}
\STATE Initialize a collection of reaction pathways $\mathcal{C} \gets \emptyset$.
\FOR{$n=1,\ldots, N$}
\STATE Sample a molecule $t \sim \mathcal{D}_{\mathtt{target}}$.
\STATE Compute a reaction pathway $\tau \leftarrow \mathcal{A}(t, p_{b})$.
\FOR{$(m, \mathcal{R})\in \tau$}
\IF{$p_{b}(\mathcal{R} | m; \bar{\theta}_{b}) > \epsilon$}
\STATE Update the collection $\mathcal{C}\leftarrow \mathcal{C} \cup \{ (m, \mathcal{R}) \}$.
\ENDIF
\ENDFOR
\ENDFOR
\STATE Initialize a collection of reaction pathways $\mathcal{C}^{\prime} \leftarrow \emptyset$.
\FOR{$(m, \mathcal{R}) \in \mathcal{C}$}
\STATE Compute $m^{\prime}\leftarrow \argmax_{m} p_{f}(m |\mathcal{R}; \theta_{f})$.
\IF{$p_{f} (m' | \mathcal{R}; \theta_{f}) > \epsilon_{\mathtt{aug}} $ and \\ $\mathcal{R} = \arg\max_{\mathcal{R}}p_{b}(\mathcal{R} | m^{\prime}; \bar{\theta}_{b})$} \STATE Update the collection $\mathcal{C}^{\prime}\leftarrow \mathcal{C}^{\prime}\cup\{(m^{\prime}, \mathcal{R})\}$.
\ENDIF
\ENDFOR
\STATE Maximize $\sum_{(m,\mathcal{R}) \in \mathcal{C} \cup \mathcal{C}^{\prime}} \log p_{b}(\mathcal{R}|m ; \theta_{b})$ over $\theta_{b}$.
   \ENDFOR
\end{algorithmic}
\end{algorithm}
To speed up training, we initialize the backward reaction model $p_{b}(m|\mathcal{R}; \theta_{b})$ with supervised learning on the reaction dataset $\mathcal{D}_{\mathtt{reaction}}$. Intuitively, our algorithm allows training the backward reaction model on reactions with their quality improved by using the search algorithm, i.e., \textsc{Retro*}. Since we additionally filter out unrealistic reactions based on the reference backward reaction model, our backward reaction model retains the ability to generate realistic reactions. 

Our self-improving procedure is similar to the prior works such as DAgger \citep{ross2011reduction}, retrospective imitation learning \citep{song2018learning}, expert iteration \cite{anthony2017thinking, anthony2019policy}, AlphaGo Zero \cite{silver2017mastering}, and NEXT \citep{Chen2020Learning} in meta path planning. Meanwhile, one can note that RetroGNN \citep{liu2020retrognn} uses routes found by a retrosynthetic planning algorithm to learn a value function, whereas our framework learns a policy (backward reaction model).

We also note that \citet{schreck2019learning} proposed an end-to-end framework based on reinforcement learning of the backward reaction model towards maximizing the success rate of finding a reaction pathway for the target molecule. However, they do not consider whether if the models propose realistic reaction pathways and are not directly comparable with our work.

We provide an illustration and a detailed description of our framework in Figure \ref{fig:concept} and Algorithm \ref{alg:algo}, respectively.

\subsection{Detailed Components of Self-Improved Retrosynthetic Planning}
In the rest of this section, we provide a detailed description of our algorithmic components: generating reaction pathways using a search algorithm, evaluating realistic-ness of reactions using a reference backward reaction model, and augmenting reactions using a forward reaction model.

\textbf{Generating reaction pathways.}
To generate reaction pathways, we sample a target molecule $t$ from the target molecule dataset $\mathcal{D}_{\mathtt{target}}$ and apply the search algorithm $\mathcal{A}$ based on the current backward reaction model $p_{b}(\mathcal{R} | m; \theta_{b})$. In this work, we consider \textsc{Retro*} \citep{chen2020retro} for generating the reaction pathways. However, our framework is general and applicable to other existing search algorithms such as Monte Carlo tree search \cite{segler2018planning} and proof number search \cite{kishimoto2019depth}.


\textbf{Filtering out unrealistic reactions.}
To prevent the backward reaction model from learning to predict unrealistic reactions, we use a reference backward reaction model $p_{b}(\mathcal{R}|m; \bar{\theta}_{b})$ to determine whether a reaction is realistic or not. To be specific, we train the reference backward model on a real-world reaction dataset $\mathcal{D}_{\mathtt{reaction}}$ and use its likelihood to measure the realistic-ness of the reactions; we filter out generated reactions whose likelihood under the reference backward model is less than $\epsilon$.


\begin{figure}[t]
\centering
\begin{subfigure}{0.9444444444\columnwidth}
\includegraphics[width=1\linewidth]{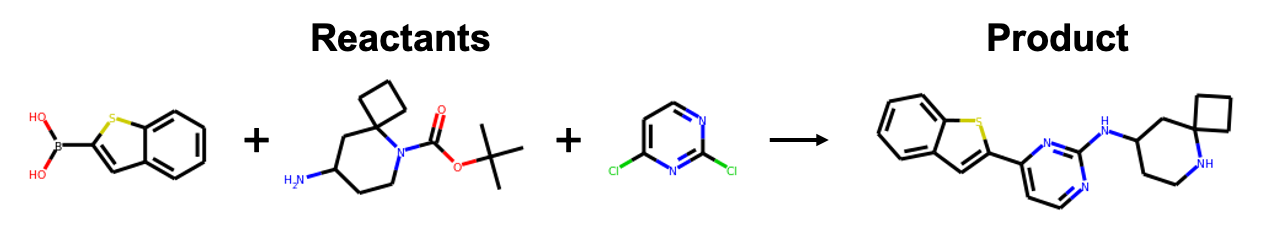}
\caption{Extracted reaction from reaction pathways}
\end{subfigure}
\begin{subfigure}{\columnwidth}
\includegraphics[width=1\linewidth]{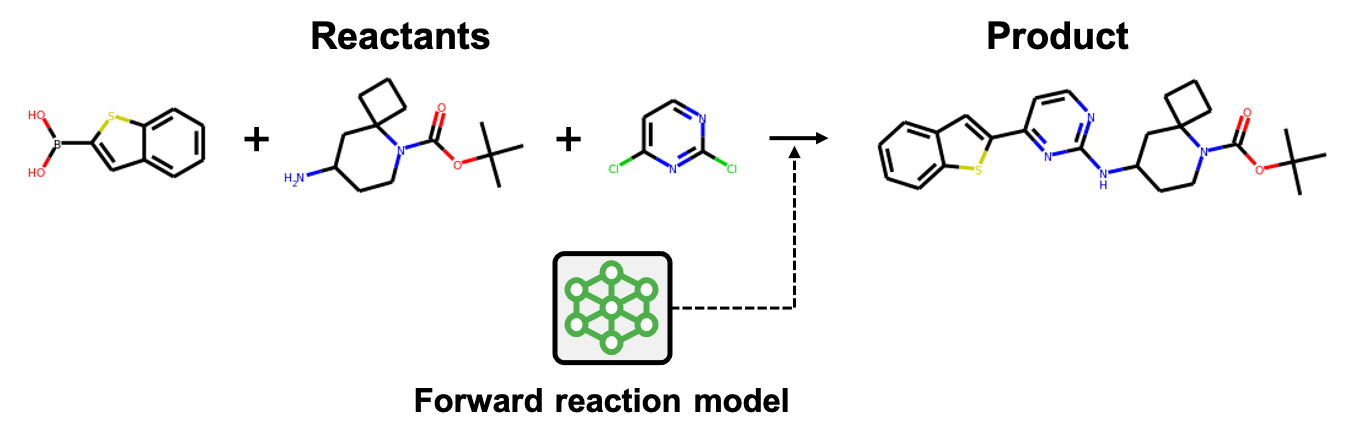}
\caption{Augmented reaction}
\end{subfigure}
\caption{Example of (a) extracted reaction from reaction pathways, and (b) corresponding augmented reaction via the forward reaction model.}
\label{fig:aug_ex}
\end{figure}
\textbf{Reaction augmentation.} The generalization capability of the backward reaction model can be improved by training on a diverse set of reactions. To this end, we propose a new augmentation scheme based on domain knowledge of chemistry; there can be multiple products resulting from the same reactant-set. Based on this knowledge, we augment the existing reaction $R=(m, \mathcal{R})$ by replacing the product $m$ with a new product $m^\prime$ proposed by a forward reaction model $p_{f}(\cdot | \mathcal{R}; \theta_{f})$. 
In order to augment the reactions in a realistic way, we additionally filter out the proposed product $m^\prime$ when it is not confident from the forward reaction model or the reference backward reaction model, i.e., we reject $m^\prime$ when $p_{f} (m' | \mathcal{R}; \theta_{f}) \leq \epsilon_{\mathtt{aug}}$ or $\mathcal{R} \neq \argmax_{\mathcal{R}} p_{b}(\mathcal{R} | m'; \bar{\theta}_{b})$. We note that our augmentation method is computationally cheaper than generating reactions from scratch. We demonstrate an example of outcomes from our augmentation scheme in Figure \ref{fig:aug_ex}. 

\section{Experiments}

\subsection{Experimental Setup}




Our framework requires specifying 
(1) a set of building block molecules $\mathcal{I}$, (2) a target molecule dataset $\mathcal{D}_{\mathtt{target}}$, (3) a reaction dataset $\mathcal{D}_{\mathtt{reaction}}$, (4) a retrosynthetic planning algorithm $\mathcal{A}$, and (5) a backward reaction model $p_{b}$, a reference backward reaction model $\bar{p}_{b}$, a forward reaction model $p_{f}$. The details of the components are as follows.

\textbf{Dataset.}
For building blocks $\mathcal{I}$, we use all of 231M commercially available molecules present in \emph{eMolecules}.\footnote{\url{http://downloads.emolecules.com/free/2019-11-01/}} Note that chemists can choose another set of molecules for $\mathcal{I}$ depending on their own circumstances, such as a financial budget. For the target molecules $\mathcal{D}_\mathtt{target}$, we choose synthesizable molecules from $\mathcal{I}$ and reactions in the United States Patent Office (USPTO) database \citep{lowe2012extraction}. To this end, we follow the procedure described by \citet{chen2020retro} and then obtain 299202 target molecules. For the reaction dataset $\mathcal{D}_{\mathtt{reaction}}$, we use reactions extracted from USPTO, following training/validation/test splits by \citet{chen2020retro}.

\textbf{Retrosynthetic planning algorithm.}
We use \textsc{Retro*} and \textsc{Retro*-0} \citep{chen2020retro} as the search algorithm $\mathcal{A}$ in our framework. \textsc{Retro*-0} denotes its variant not relying on the value function, which estimates the cost to synthesize a given molecule, proposed by \citet{chen2020retro}. 

\textbf{Model.}
We use template-based 2-layer MLPs \cite{segler2017neural} for parameterizing backward reaction model $p_{b}$, reference backward reaction model $\bar{p}_{b}$, and forward reaction model $p_{f}$. More specifically, the models predict a plausible reaction template among a pre-defined set of templates $\mathcal{T}$, i.e., they can be considered as classification models. 
We use RDChiral \citep{coley2019rdchiral} for extracting reaction templates from the USPTO database, and it results in 380k reaction templates. We use the Morgan fingerprint \citep{rogers2010extended}, of radius 2 with 2048 bits, as an input of MLPs.

\begin{table*}[ht!]
\caption{Performance of backward reaction model and retrosynthetic planning in the USPTO dataset. The backward reaction model is evaluated using \textsc{Top-1} and \textsc{Top-10} exact match accuracy (\%). Retrosynthetic planning is evaluated using \textsc{Succ. Rate $(N=50)$}, \textsc{Succ. Rate $(N=500)$}, \textsc{Length}, \textsc{Time}, and \textsc{Cost}. $N$ denotes the limit of backward reaction model calls. \textsc{Length} is the number of reactions in a route. \textsc{Time} is measured by the number of backward reaction model calls, with a hard limit of 500. $^{\dagger}$The experimental results of \textsc{Greedy DFS}, \textsc{MCTS}, and \textsc{DFPN-E} are from \citet{chen2020retro}.  The best results are marked in bold. We use brackets to report the relative gains over each counterpart that does not use our framework.}
\label{tab:1_main}
\vspace{-0.1in}
\begin{center}
\begin{sc}
\resizebox{\textwidth}{!}{
\begin{tabular}{l cc ccccc}
\toprule
& \multicolumn{2}{c}{Reactions} & \multicolumn{5}{c}{Reaction pathways} \\
\cmidrule(lr){2-3} \cmidrule(lr){4-8}
Algorithm & Top-1 $\uparrow$ & Top-10 $\uparrow$ &\thead{Succ. rate $\uparrow$ \\ $(N=50)$} & \thead{Succ. rate $\uparrow$ \\ $(N=500)$} & Length $\downarrow$ & Time $\downarrow$ & Cost $\downarrow$ \\
\midrule
Greedy DFS$^{\dagger}$ & - & - & - & 22.63 & - & 388.15  &  - \\
MCTS$^{\dagger}$ & - & - & - & 33.68 & - & 370.51  &  - \\
DFPN-E$^{\dagger}$ & - & - & - & 55.26 & - & 279.67  &  - \\
\midrule
Retro*-0 & \textbf{44.53} & 72.71 & 27.37 & 79.47 & 11.21 & 208.09  &  19.40 \\
\rowcolor{Gray}Retro*-0 + ours      & 44.03 & 73.14      & 57.37   & \textbf{96.32} &  \textbf{7.69} & \textbf{96.22}  &  \textbf{11.66} \\
\rowcolor{Gray} & (-1.12\%) & (+0.59\%) & (+109.62\%) & (+21.20\%) & (-31.40\%) & (-53.76\%) & (-39.90\%) \\
\midrule
Retro* & \textbf{44.53} & 72.71 & 44.21 & 86.84 & 9.71 & 157.11  &  15.33 \\
\rowcolor{Gray}Retro* + ours        & 44.03 & \textbf{73.15} & \textbf{57.89} & 91.05 & 8.74 & 100.15  &  15.23 \\
\rowcolor{Gray} & (-1.12\%) & (+0.61\%) & (+30.94\%) & (+4.85\%) & (-9.99\%) & (-36.25\%) & (-0.65\%) \\
\bottomrule
\end{tabular}}
\end{sc}
\end{center}
\vskip -0.1in
\end{table*}
\begin{figure}[t]
\vspace{0.1in}
    \centering
    \includegraphics[width=0.45\textwidth]{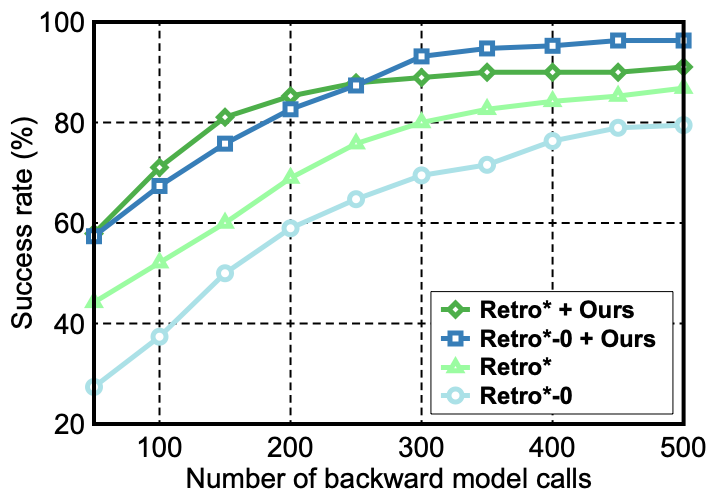}
    \caption{Success rate (\%) under varying limits of backward reaction model calls. Our framework outperforms the best baselines, \textsc{Retro*-0}, regardless of the limits.}
    \label{fig:main_graph}
\end{figure}
\begin{table}[t]
\caption{Ablation study on augmentation via the forward reaction model. \textsc{Search} indicates the type of search algorithm $\mathcal{A}$. \textsc{Aug} indicates the augmentation via the forward reaction model. $N$ denotes the limit of backward reaction model calls.} 
\label{tab:3_ablation_aug}
\vskip 0.1in
\begin{center}
\begin{sc}
\begin{tabular}{lc ccc}
\toprule
& 
&
\multicolumn{3}{c}{Succ. Rate $\uparrow$} \\
\cmidrule(lr){3-5}
Search & Aug & $N=50$ & $N=250$ & $N=500$ \\
\midrule
Retro*-0 &         & 46.84 & 81.05 & 92.63 \\
Retro*-0 & $\surd$ & \textbf{50.00} & \textbf{83.16} & \textbf{93.16} \\
\midrule
Retro* &         & \textbf{51.58} & 82.11 & 88.95 \\
Retro* & $\surd$ & \textbf{51.58} & \textbf{83.68} & \textbf{90.00} \\
\bottomrule
\end{tabular}
\end{sc}
\end{center}
\vskip -0.1in
\end{table}
\textbf{Training.}
The reference backward reaction model $\bar{p}_{b}$ and forward reaction model $p_{f}$ are trained using reactions in the dataset $\mathcal{D}_{\mathtt{reaction}}$ and then frozen before conducting our framework. Instead of training a reference backward reaction model from scratch, we use the backward reaction model trained by \citet{chen2020retro}\footnote{\url{https://github.com/binghong-ml/retro\_star}} for our reference backward model $\bar{p}_{b}$. The forward reaction model $p_{f}$ is trained with a learning rate of 0.001 for 100 epochs. The parameters of the backward reaction model $p_{b}$ are initialized to that of the reference backward reaction model $\bar{p}_{b}$. 
During the self-improving procedure in our framework, the backward reaction model $p_{b}$ is trained with a learning rate of 0.0001 for 20 epochs.
Adam optimizer \citep{kingma2014adam} is used with a mini-batch of size 1024 for training all the models. 
We iterate our overall procedure three times.

\textbf{Filtering thresholds.}
In our framework, there exists two thresholds: (1) $\epsilon$ for removing unrealistic reactions from success routes via a reference backward reaction model, (2) $\epsilon_{\mathtt{aug}}$ for rejecting unconfident augmented reaction generated by a forward reaction model. We set both thresholds $\epsilon, \epsilon_{\mathtt{aug}}$ as 0.8. Namely, we filter out reactions of which likelihood under the corresponding model is under 0.8.

\begin{figure*}[t]
\vspace{0.1in}
\centering
\begin{subfigure}{0.45\textwidth}
\includegraphics[width=0.95\linewidth]{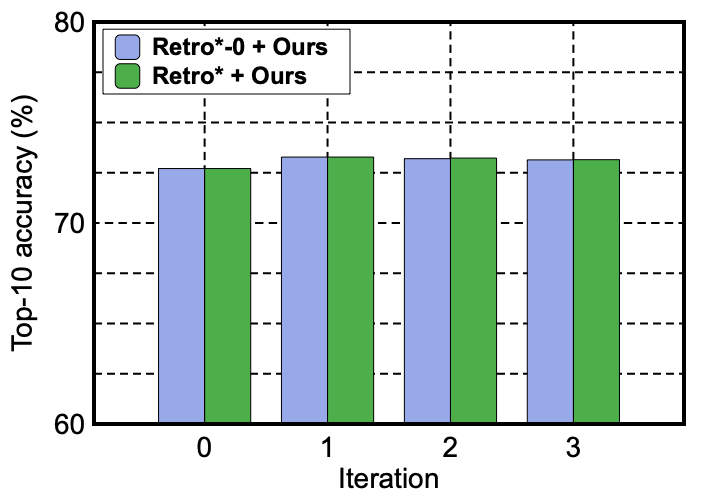}
\caption{\textsc{Top-10} accuracy under multiple iteration}
\end{subfigure}
\begin{subfigure}{0.45\textwidth}
\includegraphics[width=0.95\linewidth]{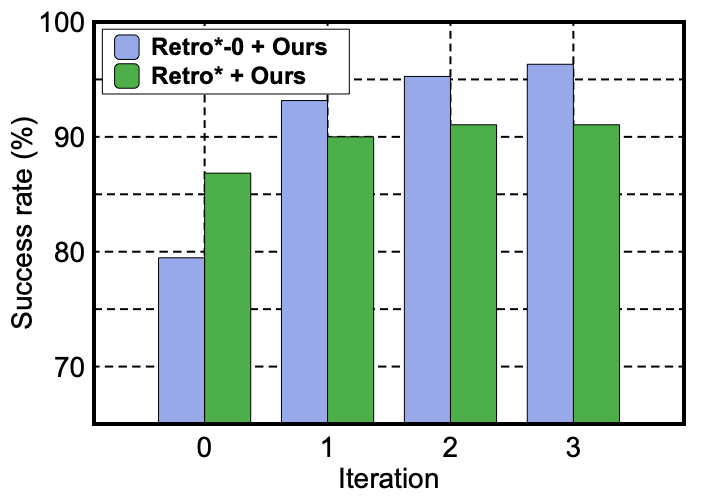}
\caption{Success rate under multiple iteration}
\end{subfigure}
\caption{We repeat our procedure multiple times and investigate the performance of the backward reaction model and the retrosynthetic planning. Iteration 0 is vanilla \textsc{Retro*-0} or \textsc{Retro*}, which our framework is not applied yet. As we iterate our framework, we can further improve the performance of retrosynthetic planning while maintaining the reliability of the backward reaction model.}
\label{fig:iter}
\end{figure*}
\begin{table*}[ht!]
\caption{Experimental results of our framework with different filtering thresholds $\epsilon$ in reaction extraction step using the reference backward model. As we do not filter out any reactions from success routes, we suffer performance degradation in \textsc{Top-1} and \textsc{Top-10} accuracy, as unrealistic reactions can be included in the success routes. If we filter out unrealistic reactions using log-likelihood under the reference backward model, we can improve the performance of retrosynthetic planning while maintaining that of the backward reaction model. We report mean and standard deviation across five independent runs.}
\label{tab:2_ablation_delta}
\begin{center}
\begin{sc}

\begin{tabular}{l cc ccccc}
\toprule
& \multicolumn{2}{c}{Reactions} & \multicolumn{5}{c}{Reaction pathways} \\
\cmidrule(lr){2-3} \cmidrule(lr){4-8}
\thead{Filtering \\ Threshold $\epsilon$} & Top-1 $\uparrow$ & Top-10 $\uparrow$ &\thead{Succ. rate $\uparrow$ \\ ($N=50$)} & \thead{Succ. rate $\uparrow$ \\ ($N=500$)} & Length $\downarrow$ & Time $\downarrow$ & Cost $\downarrow$\\
\midrule
0.9  & 44.45 \stdv{0.01} & 73.29 \stdv{0.01} & 48.42 \stdv{0.88} & \textbf{92.21} \stdv{0.70} & 8.77 \stdv{0.11} & 130.66 \stdv{1.69} & 13.23 \stdv{0.40} \\
0.8  & \textbf{44.52} \stdv{0.01} & \textbf{73.30} \stdv{0.01} & 47.68 \stdv{1.23} & 92.00 \stdv{0.21} & \textbf{8.74} \stdv{0.08} & \textbf{129.70} \stdv{2.15} & \textbf{13.01} \stdv{0.18} \\
0.7  & 44.50 \stdv{0.02} & 73.28 \stdv{0.01} & 45.79 \stdv{0.58} & 91.47 \stdv{0.84}  & 8.82 \stdv{0.21} & 131.91 \stdv{1.84} & 13.15 \stdv{0.38} \\
0.6  & 44.48 \stdv{0.01} & 73.28 \stdv{0.01} & 44.11 \stdv{0.52} & 90.95 \stdv{1.02}  & 8.92 \stdv{0.28} & 130.14 \stdv{1.02} & 13.55 \stdv{0.62} \\
0.5  & 44.46 \stdv{0.01} & 73.26 \stdv{0.00} & 43.79 \stdv{0.61} & 91.05 \stdv{0.67}  & 8.95 \stdv{0.20} & 130.39 \stdv{2.69} & 13.24 \stdv{0.43} \\
0    & 41.04 \stdv{0.02} & 71.94 \stdv{0.01} & \textbf{54.32} \stdv{0.39} & 83.58 \stdv{0.39}  & 9.91 \stdv{0.09} & 149.16 \stdv{1.90} & 18.23 \stdv{0.18} \\
\bottomrule
\end{tabular}

\end{sc}
\end{center}
\vskip -0.1in
\end{table*}
\textbf{Evaluation.}
We evaluate retrosynthetic planning for 190 target molecules in a limited time budget (i.e., the number of calls of backward reaction model $p_b$) following \citet{chen2020retro}. In the limit, we measure the success rate, the average time of planning, the average length, and the cost of discovered routes. 
Note that we consider the cost of a route as the summation of negative log-likelihoods of reactions in the route following \citet{chen2020retro}, i.e., $-\sum_{(m, \mathcal{R}) \in \tau}\log p_{b}(\mathcal{R} | m ; \bar{\theta}_{b})$.
We evaluate the backward reaction model using widely-used top-$k$ exact match accuracy in the test split of the reaction dataset $\mathcal{D}_{\mathtt{reaction}}$. 

\textbf{Baselines.}
To demonstrate the effectiveness of our framework, we compare our method with existing retrosynthetic planning frameworks such as \textsc{Greedy DFS}, \textsc{MCTS} \cite{segler2018planning}, \textsc{DFPN-E} \cite{kishimoto2019depth}, \textsc{Retro*}, and \textsc{Retro*-0} \cite{chen2020retro}. We note that the considered baselines focus on designing efficient search algorithms for retrosynthetic planning, while our main contribution is the training of the backward reaction model towards maximizing the performance of retrosynthetic planning. 



\subsection{Main Result}
We report the results for evaluating the performance of our framework and baselines in Table \ref{tab:1_main} and Figure \ref{fig:main_graph}. We observe that combining our framework with \textsc{Retro*} and \textsc{Retro*-0} significantly outperforms the baselines, including the original \textsc{Retro*} and \textsc{Retro*-0}. For example, \textsc{Retro*-0+ours} achieves the success rate of $96.32\%$ with a computation limit of $N=500$, while \textsc{Retro*} achieves $86.84\%$ as the best baseline. We also observe that \textsc{Retro*-0+ours} and \textsc{Retro*+ours} outperforms \textsc{Retro*-0} and \textsc{Retro*} significantly in terms of other evaluation metrics, e.g., the length and the cost of discovered reaction pathways.\footnote{If a reaction pathway is failed to be found, the length and the cost are set to two times of maximum length and cost among the ground-truth pathways of $\mathcal{D}_{\mathtt{target}}$, respectively, and the time is set to the limit of backward reaction model calls, i.e., 500.} Such a result demonstrates the effectiveness of our framework for retrosynthetic planning. 

\begin{figure*}[t!]
\vspace{0.1in}
\centering
\begin{subfigure}{0.44730671739\textwidth}
\centering
\includegraphics[width=0.9\linewidth]{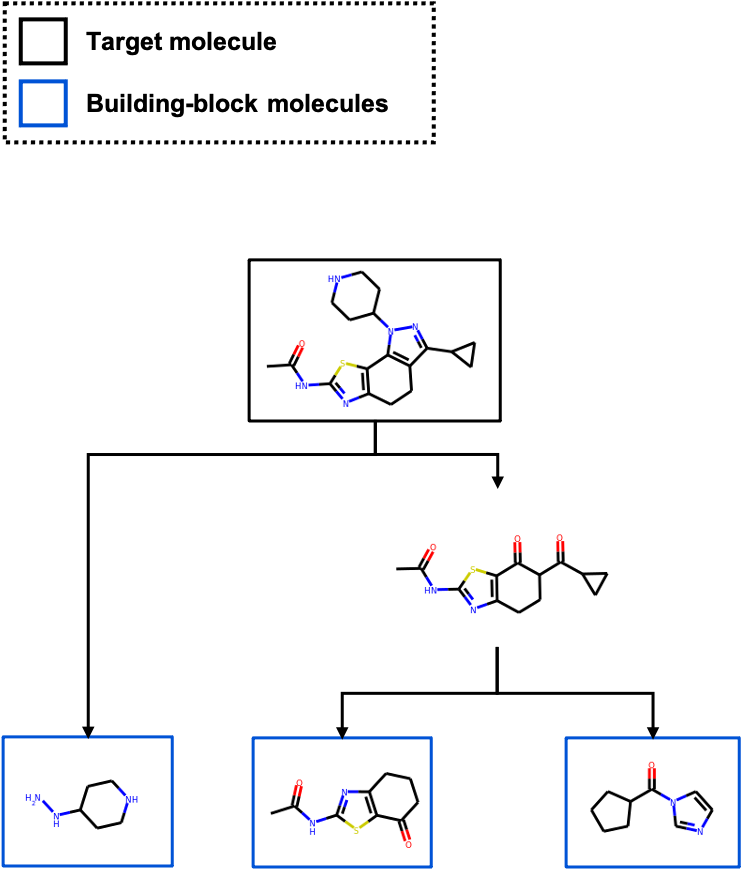}
\caption{Reaction pathway from \textsc{Retro*-0 + Ours}}
\end{subfigure}
\begin{subfigure}{0.50269328261\textwidth}
\centering
\includegraphics[width=0.9\linewidth]{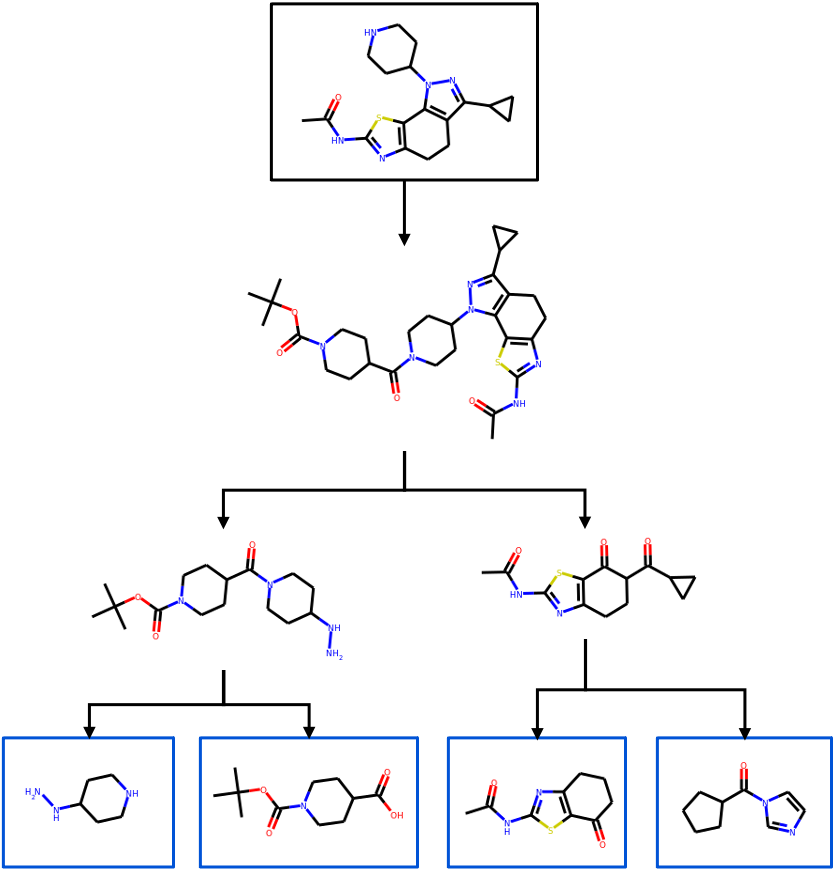}
\caption{Reaction pathway from \textsc{Retro*-0}}
\end{subfigure}
\caption{Reaction pathways from the same target molecule, 
searched by (a) \textsc{Retro*-0 + Ours} and (b) \textsc{Retro*-0}. In the example, our reaction pathway has a shorter length, which implies that our solution has better quality than that from \textsc{Retro*-0}, as shorter reaction pathways are easier to be conducted in laboratories.}
\label{fig:case_study}
\end{figure*}
On the other hand, our backward reaction model does not suffer from a drop in \textsc{Top-$k$} accuracy. Instead, they even outperform the backward reaction model trained on the reaction dataset $\mathcal{D}_{\mathtt{reaction}}$ via supervised learning in terms of \textsc{Top-10} accuracy. 
We understand that such an improvement in the \textsc{Top-10} accuracy comes from "diverse" solution candidates generated by our backward reaction model, which is encouraged by being trained on a large variety of samples, e.g., augmented reactions.
Intriguingly, we also observe that \textsc{Retro*-0+ours} performs similarly with \textsc{Retro*+ours} without the help of an additional value function for guiding the search algorithm, i.e., \textsc{Retro*}. We hypothesize that the performance of the search algorithm may saturate when the backward reaction model appropriately adapts to the search algorithm. This highlights the importance of training an appropriate backward reaction model for retrosynthetic planning.

\subsection{Ablation Study}

Next, we conduct ablation studies on our framework to investigate the effect of (1) removing the reaction augmentation procedure, (2) varying the number of iterations, and (3) varying the filtering threshold $\epsilon$ for realistic reactions. 
For the ablation studies on (1) and (3), our overall procedure is conducted a single time, i.e., we do not iterate our framework multiple times.

\textbf{Effectiveness of reaction augmentation.}
To evaluate whether if the proposed reaction augmentation is effective, we evaluate our framework without the reaction augmentation scheme. As shown in Table \ref{tab:3_ablation_aug}, our framework achieves a higher success rate for finding the reaction pathway for the target molecules. This validates the effectiveness of our reaction augmentation scheme based on the forward reaction model. 


\textbf{Number of iterations.} 
In Figure \ref{fig:iter}, we report the performance of the backward reaction model and the success rate for finding the reaction pathways over multiple iterations. The result demonstrates that iterating our framework improves the success rate of finding a reaction pathway while maintaining the accuracy of the backward reaction model. This indeed validates that our framework is effective without compromise in the ability to model realistic reactions.


\textbf{Filtering threshold $\epsilon$.}
\label{ablation:filter}
To recognize the effectiveness of filtering out unrealistic reactions within success routes, we compare the performance of our framework with varying thresholds for filtering out reactions.\footnote{In this experiment, we do not include the augmented reaction data via the forward reaction model.} 
As shown in Table \ref{tab:2_ablation_delta}, the performance of our backward reaction model is (a) improved when using the threshold, i.e., using Thr$>$0 instead of Thr$=$0, and (b) robust to the change of hyper-parameter, i.e., varying the threshold from $\{0.5, 0.6, 0.7, 0.8, 0.9\}$. In particular, setting \textsc{Filtering Threshold} $\epsilon$ to 0, the top-1 exact match accuracy of our backward reaction model degrades from 44.52\% to 41.04\%. On the other hand, if we filter out unrealistic reactions by the reference backward reaction model, our backward reaction model retains its reliability and enjoys the improved performance in retrosynthetic planning without compromise.

\subsection{Case Study}
In Figure \ref{fig:case_study}, we compare reaction pathways found by \textsc{Retro*-0+ours} and \textsc{Retro*-0} for the same target molecule. In the example, we observe that both frameworks can find realistic reaction pathways for the given target molecule. However, the reaction pathway searched by \textsc{Retro*-0+ours} is preferable according to our predefined criteria, i.e., the number of reactions required for execution.

\section{Conclusion}
We propose a new framework based on self-improved model adaptation to improve retrosynthetic planning. Our main idea is to train a backward reaction model to imitate success routes found by the retrosynthetic planning algorithm combined with itself. We also propose an additional augmentation scheme that diversifies training through generating new reactions from a forward reaction model. Experiments show that our framework successfully adapts the backward reaction model to generate reactions that are both realistic and executable from building block molecules.
Meanwhile, improving filtering modules using negative (unrealistic) reaction samples and developing better reaction augmentation schemes would be interesting directions to explore.

\section*{Acknowledgements}
We thank Yeonghun Kang, Seonyul Kim, and Sangwoo Mo for providing helpful feedback and suggestions in preparing the early version of the manuscript.
We would like to thank Binghong Chen for providing the dataset and source implementation of \textsc{Retro*}.
This work was supported by Institute of Information \& Communications Technology Planning \& Evaluation (IITP) grant funded by the Korea government (MSIT) (No.2019-0-00075, Artificial Intelligence Graduate School Program (KAIST)).
This work was supported by Institute for Information \& communications Technology Planning \& Evaluation(IITP) grant funded by the Korea government(MSIT) (No. 2019-0-01396, Development of framework for analyzing, detecting, mitigating of bias in AI model and training data).

\bibliography{reference}
\bibliographystyle{icml2021}

\newpage
\onecolumn
\appendix
\section{More Discussion on Reaction Pathways}
\begin{figure}[h!]
\vspace{0.1in}
\centering
\includegraphics[width=0.45\linewidth]{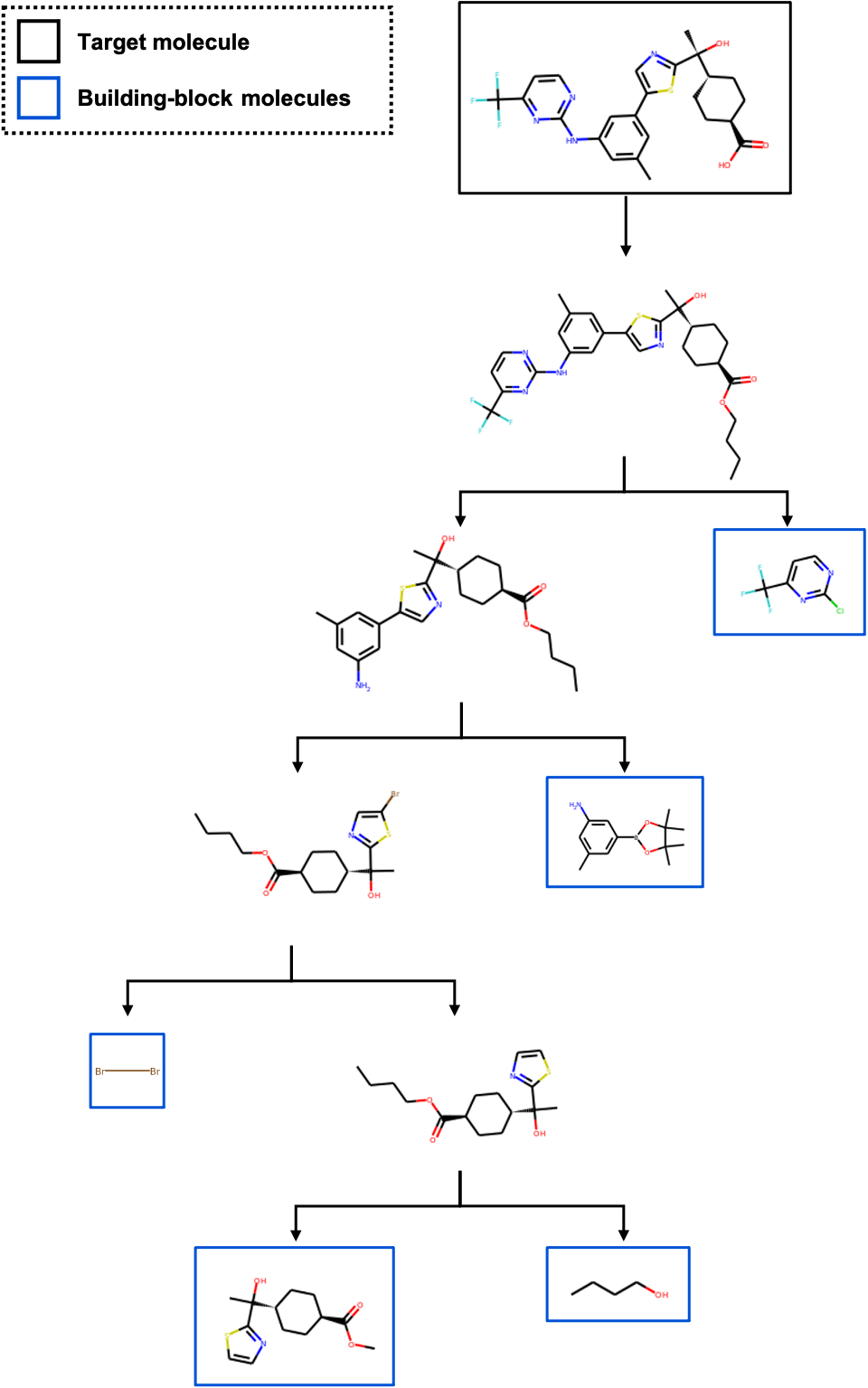}
\caption{Reaction pathway produced by \textsc{Retro*-0 + ours}, given the target molecule trans-4-{(1S)-1-Hydroxy-1-[5-(3-methyl-5-{[4-(trifluoromethyl)pyrimidin-2-yl]amino}phenyl)-1,3-thiazol-2-yl]ethyl}cyclohexanecarboxylic acid, where \textsc{Retro*-0} failed to find corresponding reaction pathway.}
\label{fig:appendix_a1}
\end{figure}

\begin{figure*}[h!]
\vspace{0.1in}
\centering
\begin{subfigure}{0.57\textwidth}
\centering
\includegraphics[width=0.9\linewidth]{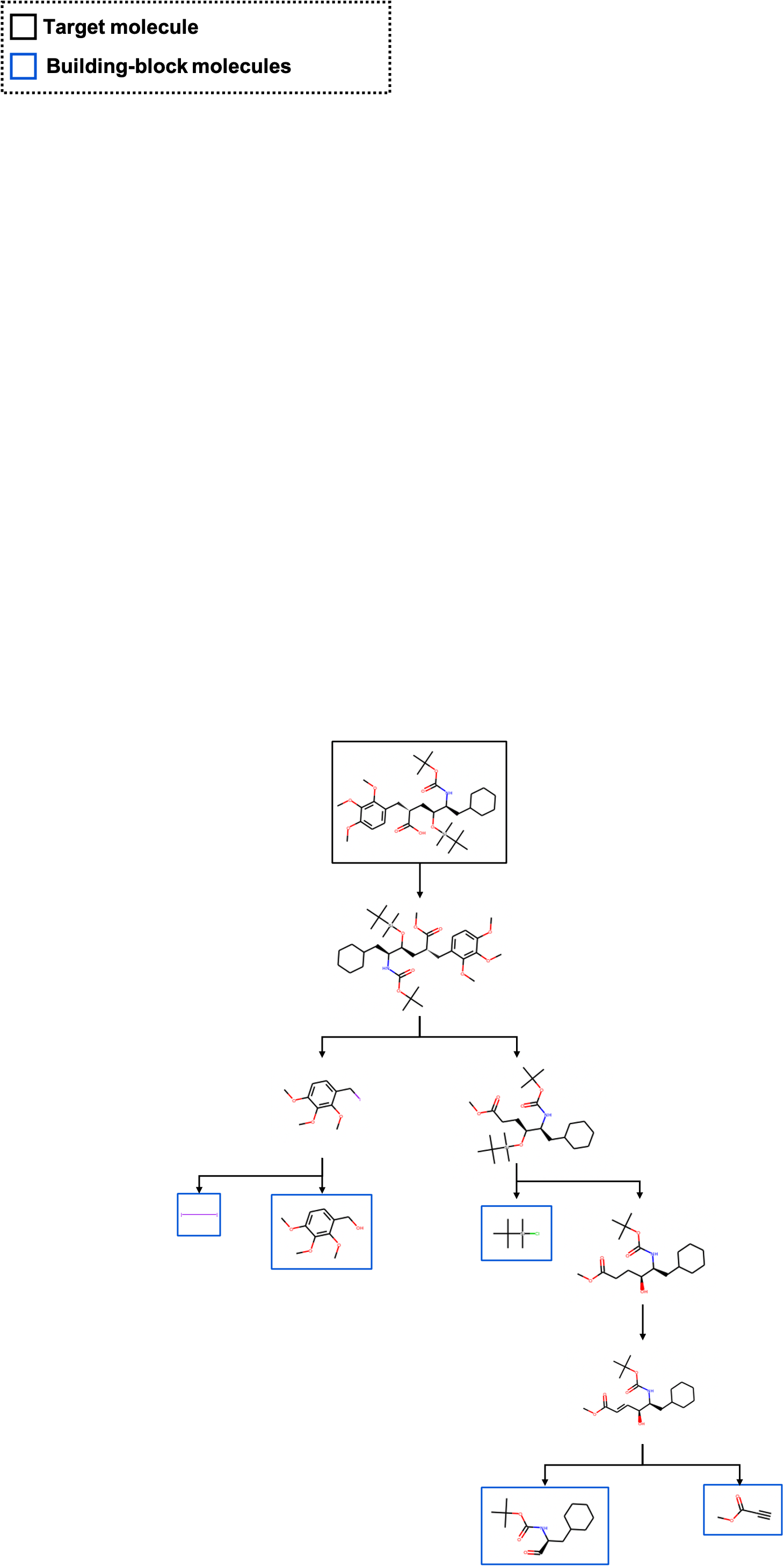}
\caption{Reaction pathway searched by \textsc{Retro*-0 + ours}}
\end{subfigure}
\begin{subfigure}{0.4\textwidth}
\centering
\includegraphics[width=0.9\linewidth]{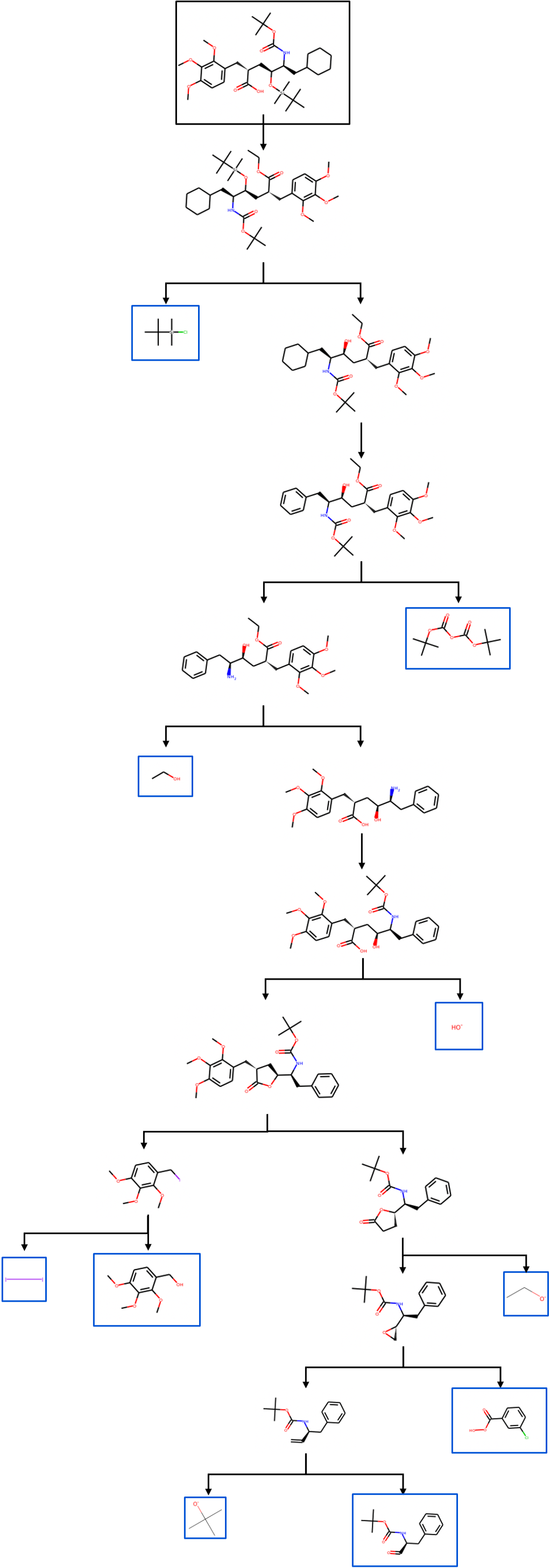}
\caption{Reaction pathway searched by \textsc{Retro*-0}}
\end{subfigure}
\caption{Reaction pathways from the same target molecule, 5(S)-(Boc-Amino)-4(S)-(tert-butyldimethylsilyloxy)-6-cyclohexyl-2(R)-[(2,3,4-trimethoxyphenyl)-methyl]hexanoic acid, searched by (a) \textsc{Retro*-0 + ours} and (b) \textsc{Retro*-0}. Our framework searches shorter reaction pathway, which is preferable in laboratories.}
\label{fig:appendix_a2}
\end{figure*}

In this section, we provide additional reaction pathways generated by our framework and baselines. In Figure \ref{fig:appendix_a1}, given the target molecule, \textsc{Retro*-0 + ours} successfully finds the reaction pathway while \textsc{Retro*-0} cannot find anyone. In Figure \ref{fig:appendix_a2}, given another target molecule, both \textsc{Retro*-0 + ours} and \textsc{Retro*-0} successfully find reaction pathways. However, \textsc{Retro*-0 + ours} finds a much shorter reaction pathway, which is preferable in laboratories. As our backward reaction model is trained to consider executability from building blocks as well as realistic-ness, our framework is able to search shorter reaction pathways.

\clearpage
\section{More Examples of Reaction Augmentation}
\begin{figure*}[h!]
\vspace{0.1in}
\centering
\begin{subfigure}{0.45\textwidth}
\centering
\includegraphics[width=0.9\linewidth]{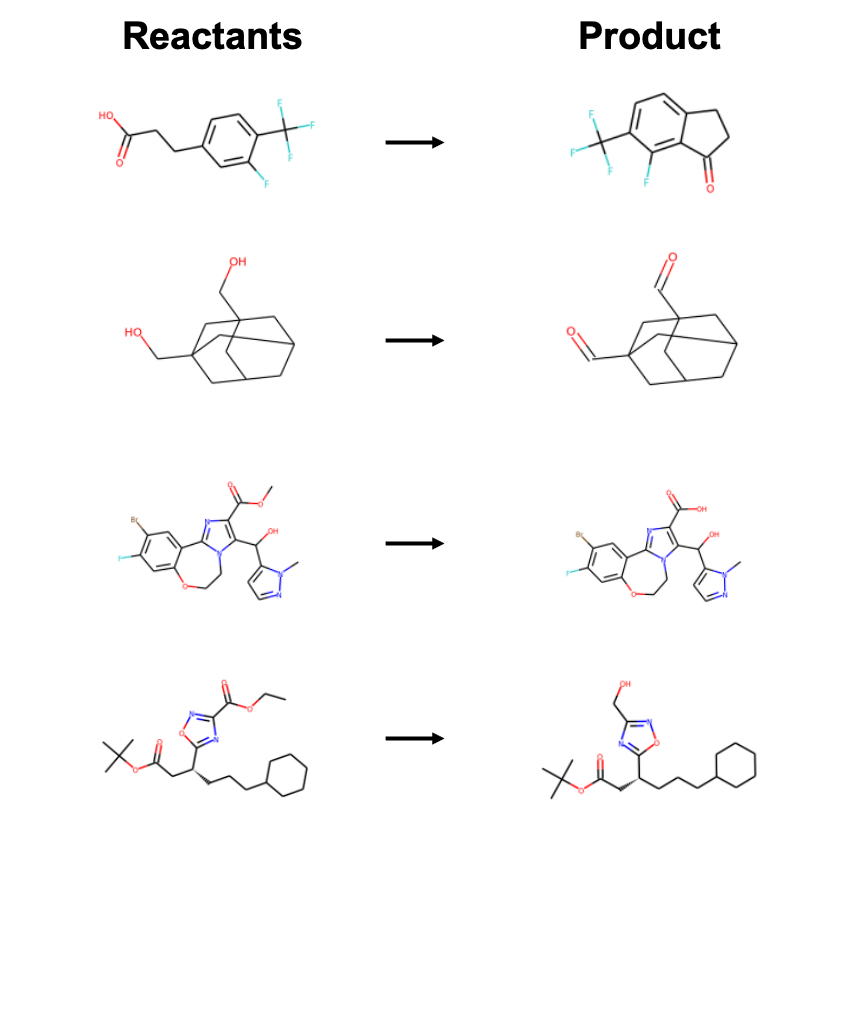}
\caption{Extracted reactions from reaction pathways}
\end{subfigure}
\begin{subfigure}{0.45\textwidth}
\centering
\includegraphics[width=0.9\linewidth]{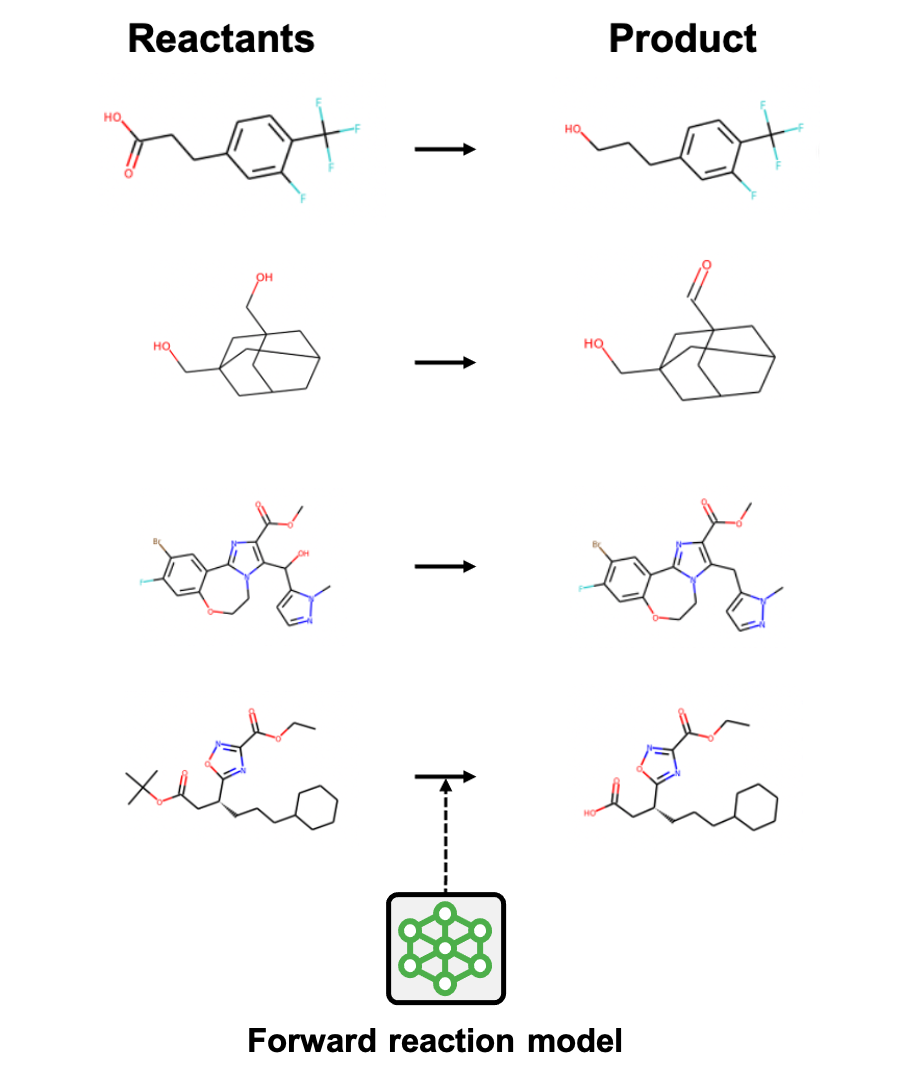}
\caption{Augmented reactions}
\end{subfigure}
\caption{Illustration of reactions (a) extracted from reaction pathways found by search algorithm combined with the backward reaction model and (b) corresponding augmented reactions via the forward reaction model. The products of augmented reactions have similar but different structures from the original products. Using augmentation, we can improve the generalization ability of the backward reaction model.}
\label{fig:appendix_b}
\end{figure*}

\clearpage

\section{Additional experiments}
\subsection{Trade-off of search time and performance}
\begin{figure}[h]
\centering
\includegraphics[width=0.415\linewidth]{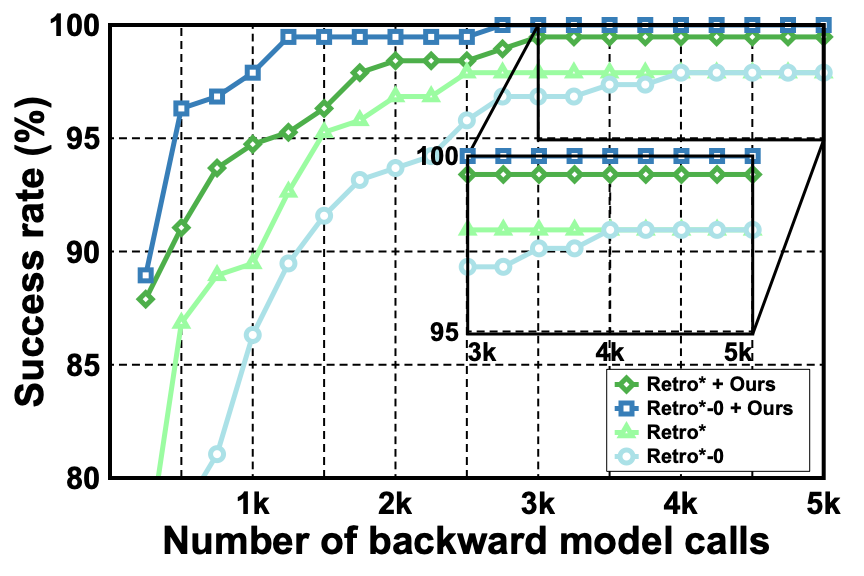}
\caption{Success rate (\%) under sufficient search time to converge, i.e., 5000 model calls. Our framework outperforms the best baselines, \textsc{Retro*-0}, even if a sufficient search time is given.}
\label{fig:suff_search}
\vspace{-.25in}
\end{figure}

We compare algorithms using a sufficient search time to converge, i.e., 5000 model calls, in Figure \ref{fig:suff_search}. Although success rates could converge if infinite search time is given, the figure highlights that (1) the gap does not converge to zero (+2.11 \%) and (2) our method can discover successful routes much efficiently. We would like to emphasize that (2) is critical in practice because the synthesis of some complex substances, i.e., vitamin B12 requires more than 100 reactions \citep{woodward1973total}, which could increase the search space exponentially.

\subsection{Comparison to another augmentation scheme}
\label{supp_mix_aug}
We compare our augmentation scheme to the ``mixed forward/reverse augmentation'' (Mix) \citep{tetko2020state}.\footnote{Note that SMILES-based augmentations, also proposed in \citep{tetko2020state}, are not directly applicable to our implementation since our model is based on Morgan fingerprint which is invariant to the SMILES-based augmentations.} The Mix augmentation augment reactions by switching products and reactants to improve generalization ability via mixed representation of latent space. As shown in Table \ref{tbl:aug_mix}, the Mix augmentation performs slightly better when using $N=50$ single-step model calls but much worse for $N=250$ and $N=500$.

\begin{table}[h]
     {
     \begin{center}{
     \resizebox{0.4\linewidth}{!}{
     \small
     \begin{tabular}{lccc}
     \toprule

     Aug & $N=50$ & $N=250$ & $N=500$ \\
     \midrule
      None     & 46.84 & 81.05 & 92.63 \\
      Mix  & \textbf{52.63}  & 66.32 & 72.63 \\
      Ours & 50.00 & \textbf{83.16} & \textbf{93.16} \\
     \bottomrule
     \end{tabular}}
     \caption{Experimental comparison between our augmentation scheme to the ``mixed forward/reverse augmentation'' (Mix) \citep{tetko2020state}. The Mix augmentation performs slightly better when using $N=50$ single-step model calls but much worse for $N=250$ and $N=500$.}
     \label{tbl:aug_mix}
     \vspace{-.2in}}
     \end{center}
     }
\end{table}

\end{document}